\AtBeginDocument{%
  \providecommand\BibTeX{{%
    \normalfont B\kern-0.5em{\scshape i\kern-0.25em b}\kern-0.8em\TeX}}}

\documentclass[10pt,twocolumn]{article} 
\usepackage{simpleConference}
\usepackage{authblk}
\usepackage{blindtext}
\RequirePackage{fix-cm}
\usepackage{graphicx}
\usepackage[square,sort,comma,numbers]{natbib}
\usepackage{tikz}
\usepackage{multirow, makecell}
\usepackage{pgfplots}
\usepackage{amsmath,amsfonts}
\usepackage{algorithmic}
\usepackage{textcomp}
\usepackage{xcolor}
\usepackage{url}
\usepackage{xcolor, soul}
\usepgfplotslibrary{fillbetween}
\sethlcolor{yellow}

\newcommand{\figref}[1]{Fig.~\ref{#1}}
\newcommand{\secref}[1]{Section~\ref{#1}}
\newcommand{\tabref}[1]{Table~\ref{#1}}

\begin{document}

\title{Real-World Graph Convolution Networks (RW-GCNs) for Action Recognition in Smart Video Surveillance}


\author[]{Justin Sanchez}
\author[]{Christopher Neff}
\author[]{Hamed Tabkhi}
\affil[]{Department of Electrical and Computer Engineering, University of North Carolina at Charlotte}
\affil[]{\textit {\{jsanch19,cneff1,htabkhiv\}@uncc.edu}}

\maketitle
\thispagestyle{empty}




\begin{abstract}
Action recognition is a key algorithmic part of emerging on-the-edge smart video surveillance and security systems. Skeleton-based action recognition is an attractive approach which, instead of using RGB pixel data, relies on human pose information to classify appropriate actions. However, existing algorithms often assume ideal conditions that are not representative of real-world limitations, such as noisy input, latency requirements, and edge resource constraints.

To address the limitations of existing approaches, this paper presents Real-World Graph Convolution Networks (RW-GCNs), an architecture-level solution for meeting the domain constraints of Real World Skeleton-based Action Recognition. Inspired by the presence of feedback connections in the human visual cortex, RW-GCNs leverage attentive feedback augmentation on existing near state-of-the-art (SotA) Spatial-Temporal Graph Convolution Networks (ST-GCNs). The ST-GCNs' design choices are derived from information theory-centric principles to address both the spatial and temporal noise typically encountered in end-to-end real-time and on-the-edge smart video systems. Our results demonstrate RW-GCNs' ability to serve these applications by achieving a new SotA accuracy on the NTU-RGB-D-120 dataset at 94.1\%, and achieving 32$\times$ less latency than baseline ST-GCN applications while still achieving 90.4\% accuracy on the Northwestern UCLA dataset in the presence of spatial keypoint noise. RW-GCNs further show system scalability by running on the $10\times$ cost effective NVIDIA Jetson Nano (as opposed to NVIDIA Xavier NX), while still maintaining a respectful range of throughput ($15.6$ to $5.5$ Actions per Second) on the resource constrained device. The code is available here: https://github.com/TeCSAR-UNCC/RW-GCN.

\end{abstract}



\section{Introduction}
\label{intro}
Action recognition, the process of identifying actions performed by actors in a video segment, is a challenging and prolific field of computer vision. This challenge is only exacerbated by the constraints of on-the-edge real-time applications. Edge computing is a requirement for many applications that aim for maintaining privacy and scalability. Recent advancements in deep neural networks (DNNs) have lead to a renaissance of the field, driving model accuracy far beyond what was previously possible. This is especially pronounced in skeleton-based action recognition, which instead of using RGB pixel data, relies on human pose information to classify appropriate actions. In particular, methods derived from graph convolution networks (GCNs) \cite{kipf2016semi}, specifically Spatial-Temporal Graph Convolution Networks \cite{yan2018spatial} (ST-GCNs), have pushed state-of-the-art (SotA) action recognition to new heights. Currently, near SotA accuracy can be achieved with a simple baseline ST-GCN algorithm, achieving up to 87\% accuracy. With accuracy this high being available even to relatively simple models, the realization of complex computer vision applications that rely on understanding the actions of people (e.g. video surveillance, autonomous vehicles, patient monitoring) is now within the realm of possibility. However, many of these applications have real-world constraints that have yet to be addressed properly, such as the need for an on-the-edge end-to-end system to meet privacy and scalability concerns and real-time latency constraints. These real-world constraints inhibit the deployment of existing skeleton-based action recognition works.

Existing works that focus on skeleton-based action recognition often assume ideal conditions that are not representative of the real-world context. Typically, the pose information used by the network is assumed to be perfect and is often based on hand-annotated data. However, in the case of a real-world system, skeletal pose information would come from an imperfect human pose estimator, leading to imperfect pose data. Even the most advanced SotA pose estimators, with their tens of millions of parameters and hundreds of billions of operations \cite{openpose2018,efficientHRNet}, cannot produce flawless pose data. The accuracy of these pose estimators is further hampered by the resource constraints of edge devices. Additionally, these works often neglect satisfying the latency constraints of these real-world applications on edge devices, which decidedly limits their suitability towards real-world use.

While most works focus primarily on achieving high accuracy, there are some that attempt to address specific aspects of real-world conditions. Specifically, noisy data and real-time latency constraints with respect to end-to-end edge systems. Noisy data can be seen as incomplete human skeletons due to scene occlusions, or missing skeletons in multi-person settings caused by ID confusion. The predominant method for emulating real-world analysis is through evaluation on the Kinetics-Skeleton dataset \cite{kay2017kinetics}. This is an augmentation of the popular Kinetics action dataset derived by using OpenPose \cite{openpose2018} to extract human pose information from the original RGB images. However, accuracies on this dataset tend to be low, with SotA being under 50\%. This can partially be explained by the noise introduced by using an imperfect pose estimator, but the primary factor is that, for this particular dataset, pose data does not include a lot of action-salient information. Contributors to this are scenes with moving cameras, interactions from a first-person view, and information dependent on the context of a scene. 

Overall, existing approaches fail to address real-world action recognition constraints including both latency constraints and noisy data. This failure denies the realization of many applications with on-the-edge and real-time requirements. This creates demand for a new design mindset, formalization, and domain specification that copes with and even addresses these challenges while maintaining robust accuracy. This work presents Real-World Graph Convolution Networks (RW-GCNs) as the solution to cope with the inherent constraints of real-world action recognition. By doing so, we are enabling the deployment of real-world action recognition on edge devices. We look toward the information bottleneck theory of deep learning \cite{tishby2015deep} and with inspiration from a neuroscience perspective \cite{kreiman2020beyond} we derive attentive feedback augmentations as solutions that simultaneously deal with latency constraints and noisy keypoints. RW-GCNs' feedback solves the issue of maintaining accuracy while operating with latency constraints on edge devices by storing and reusing past features and implicit temporal sampling of important features. The attentive feedback reexamines past features and enhances the discriminatory ability of the model. We further derive two variations of feedback for RW-GCNs: (1) semantic feedback and (2) control feedback.

For a holistic evaluation with respect to on-the-edge real-world constraints, we identify a realistic edge system for skeleton-based action recognition that satisfies privacy and scalability constraints that existing cloud based systems would fail to. Furthermore, we introduce two new computational performance metrics: Actions per Second ($ApS$), a method of analyzing throughput across multiple latency constraints, and Action Product Delay ($APD$), an end-to-end system latency metric. We then quantitatively evaluate RW-GCNs, with both semantic feedback and control feedback, to address real-world action recognition challenges under latency constraints and noisy input conditions induced by the real-world edge system. Our results show that RW-GCNs achieve a new SotA accuracy of \textbf{94.16}\% for the NTU-RGB-D-120 dataset \cite{liu2020ntu} that has \textbf{$3.02\times$} less latency over the ST-GCN baseline. Evaluating the Northwestern UCLA dataset \cite{wang2014NwUCLA} reveals that RW-GCNs can achieve \textbf{90.4\%} accuracy with \textbf{32.5}$\times$ less latency than the baseline ST-GCNs. This is despite spatial keypoint noise being present in the validation and training of the Northwestern UCLA dataset. Finally, RW-GCNs are able to run in the presence of fully end-to-end system noise with \textbf{$32.5\times$} less latency than baseline ST-GCNs while maintaining \textbf{$71.8$\%} accuracy on the Northwestern UCLA dataset.  
In summary, the proposed work makes the following contributions:
\begin{itemize}
  \item We define domain constraints and introduce new evaluation methods for the domain of real-world skeleton-based action recognition on-the-edge.
  \item Introduce RW-GCNs as a solutions to the real-world constraints of latency and noisy data simultaneously though attentive feedback architecture augmentations based on information theory.
  \item Systematically evaluate latency constraints with respect to unconstrained existing works on real edge devices.
  \item Conduct an ablation study of an end-to-end action recognition system, and the emergent system noise on a video surveillance oriented dataset.
\end{itemize}

In the following, \secref{sec:2} goes over related works. \secref{sec:3} presents our domain constraints and evaluation metrics with respect to a real-world end-to-end edge system and the emergent system noise associated with it. \secref{sec:4} proposes RW-GCNs as a solution towards enabling skeleton-based action recognition on-the-edge and in real-time, and explains the theoretical foundations and implementation details of both semantic and control feedback based implementations of RW-GCNs. \secref{sec:5} presents the results, analysis, and evaluation of this work. \secref{sec:6} contains our concluding remarks.

\footnote{This is a pre-print of an article to be published in the Sixth ACM/IEEE Symposium on Edge Computing  SEC '21, December 14–17, 2021, San Jose, CA, USA,{https://doi.org/10.1145/3453142.3491293}}

\section{Related Works}
\label{sec:2}

Action recognition can typically be done either directly from video information \cite{lin2019tsm} or by utilizing human pose information through time. The former method can be called pixel-based action recognition and has been extensively explored with large datasets \cite{kay2017kinetics}. Pixel-based action recognition approaches must be robust with respect to the spatial appearance features of individual actors, while simultaneously capturing the temporal dynamics of the actor as done in Jiang et al. \cite{jiang2019stm}. This is made more challenging due to the the noise in the spatial domain. Existing works utilize pre-training on large image datasets before moving to the action datasets as seen in Tran et al. \cite{tran2018closer}. The temporal dynamics of action recognition also need to be accounted for, as seen in Liu et al. where video segments are viewed at multiple granularities \cite{liu2019multi}. The alternative to pixel-based action recognition is skeleton-based action recognition which intuitively has a less noisy input when compared to RGB video streams. Correctly extracting that information and classifying a particular action has been an ongoing research challenge. Previous methods relied on the expressive power of Recurrent Neural Network (RNN) based models to handle the graph structured input \cite{lee2017ensemble}. Graph Convolutional Networks (GCNs) as seen in the work of Kipf et al. \cite{kipf2016semi} utilized convolution on irregular non-euclidean graphs as opposed to the regular graphs found in images. This idea was applied to human pose graphs for action recognition in the work of Yu et al. \cite{yan2018spatial}. The end result was ST-GCNs \cite{yan2018spatial}. RW-GCNs can be seen as an intermedium between RNN and ST-GCN based approaches similar to AGC-LSTMs \cite{si2019attentionLSTMGCN}.

Current SotA approaches in the field of skeleton-based action recognition are primarily derived from (or at least related to) the work of Yang et al.'s ST-GCN \cite{yan2018spatial}. These works build on top of ST-GCNs with a multitude of innovations. The work of Liu et al. \cite{liu2020disentangling} addresses the existing biases of the baseline ST-GCN formulation. The first bias is against the representation of long-range joint relationships. This bias is caused by the ST-GCN's feature aggregation being dominated by local correlations due to redundant connections in the adjacency matrix. The proposed solution is the removal of the existing redundant connections. The second bias is against the representation of complex spatial-temporal feature propagation. This is due to the factorization of spatial and temporal graph convolution into separate layers. This work additionally proposes a unified operator that concurrently processes the spatial and temporal aspects of the data.

Rather than encode hand crafted design principles into  the network, several works utilize an adaptive graph structure that allows the network to learn design principles that lead to more robust representations. The work of Plizzari et al. \cite{plizzarispatial} incorporates transformers, as seen in the natural language processing domain \cite{vaswani2017attention}. The flexibility of transformers, and ultimately their core component self-attentions, enable the ST-GCN to dynamically model joint connections and learn which connections are important dependent on the data. While Plizzari et al. \cite{plizzarispatial} utilizes self-attention to overcome the limitations of the baseline ST-GCN formulation, it still utilizes the graph convolution formation (i.e. explicit convolution over an irregular graph). The work of Shi et al. \cite{shi2020decoupled} goes beyond this and entirely replaces GCNs with decoupled self-attention (decoupled self-attention is factorized across space and time). By utilizing positional encodings this work relies on the ability of self-attention to implicitly model GCNs. As opposed to increasing the flexibility of the graph structure, some other approaches improve the input representation, by making it more informative \cite{shi2019twoStream,shi2019Directed,li2019actional}. 
These works further define the input in terms of bones and motion explicitly by preprocessing the existing joint information.

While these innovations contribute to more robust generalization and overall higher accuracy on key datasets, few attempt to account for the realistic constraints and difficulties of deploying ST-GCNs in the real world. As such, most existing works fail to enable real world deployment, especially considering the computational limitations of edge devices required for certain applications. An exception is in the work of Song et al. \cite{song2020stronger} which improves the input representation similar to \cite{shi2019twoStream,shi2019Directed,li2019actional}, and then builds an efficient architecture focused on part-wise attention. The work of Cheng et al. \cite{cheng2020skeleton} also improves real-time performance by utilizing temporal-shift-modules (TSMs) \cite{lin2019tsm} as an efficient alternative towards temporal convolution. The work of Yang et al. \cite{yang2020feedback} utilizes a feedback mechanism to increase the flexibility of the temporal sampling strategy, and to additionally decrease the number of frames needs for the classification. Lastly, the work of Yu et al. addresses the real-world effects of noisy keypoints and defines the types of noise and a solution based on training with artificial noise \cite{yu2020predictively}.

Our work goes in a different direction when considering the requirements of real-world deployment. As opposed to the work of Cheng et al. and Song et al. \cite{cheng2020skeleton,song2020stronger}, we do not make explicit contributions to efficient models. Rather we focus on the input data dependency and decouple our model from large input requirements in order to meet latency constraints, similar to Yang et al. \cite{yang2020feedback}. However, rather than utilizing a densely connected layer to control the necessary feedback, we propose a feedback mechanism based on attention mechanisms similar to those found in Plizzaris et al. \cite{plizzarispatial}. Additionally, we also consider the keypoint noise similar to what was done by Yu et al. \cite{yu2020predictively}, though we do so in the context of a realistic end-to-end action recognition system. By doing so we hope to enable skeleton-based action detection for deployment in real-time edge applications.

\section{Real-World Action Recognition: Domain Constraints, Gaps, and Methods}
\label{sec:3}

\begin{figure*}[t!]
    \centering
    \includegraphics[width=1\linewidth, trim= 18 15 18 30, clip]{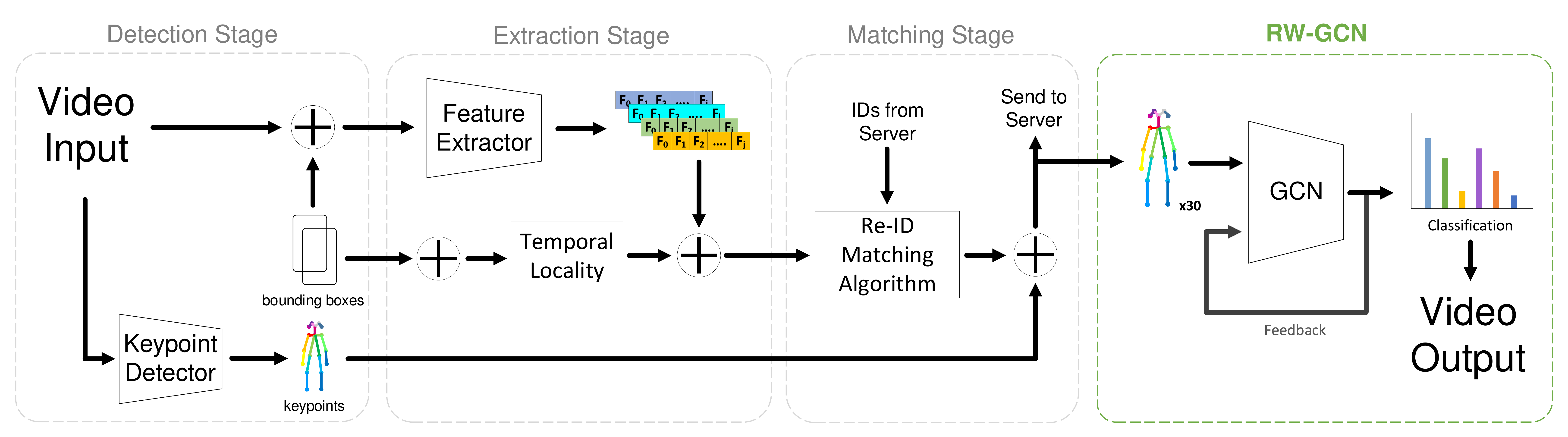}
    \caption{An example of an end-to-end system for smart video surveillance applications without using facial recognition.}
    \label{fig:end-to-end}
\end{figure*}
While the domain of skeleton-based action recognition has enjoyed increasing success, algorithmic performance with respect to real-world constraints has largely been unexplored. Our work hopes to initiate the exploration of real-world skeleton-based action recognition, which will subsequently enable skeleton-based action recognition deployment on-the-edge. In this section, we analyze the real-world deployment limitations for applications such as patient monitoring and security and define the sub-domain \textit{Real-World Skeleton Action Recognition}. At first, we briefly review real-world action recognition constraints and the necessity for an edge computing paradigm. Then we introduce two new evaluation metrics to capture the latency constraints. Finally, we dive into the details of noisy data in the context of a full end-to-end vision processing pipeline for video surveillance applications on-the-edge.


\subsection{Real-World Constraints} 

\textbf{Noisy Input:} The first primary constraint of this new sub-domain is real-world data. Skeleton-based action recognition is highly dependent on human pose data in the form of key points. Generally, works published in the field assume perfect human pose data is provided. This pristine data is often obtained through time consuming manual annotation or the use of expensive, highly accurate sensors in a laboratory environment. In real-world deployment, pose data needs to be generated on the fly, often by a dedicated machine learning algorithm, resulting in imperfect data. This is referred to as noisy data, and in this paper we assume this noise is generated by a DNN-based multi-person pose estimator. We define noise in a similar manner as in the work of Yu et al. \cite{yu2020predictively}.

We adopt their definition of spatial noise; that is human keypoint joints that should be in the frame but are not provided. However, we expand upon their definition of temporal noise, separating it into two distinct categories. The first category is frame-level temporal noise. This occurs when an individual's data is completely missing from a frame, whether due to failure to detect a person or by misidentifying them in a multi-person environment. The second category is receptive field level temporal noise. This is an implicit noise, defined as a limitation of a network's capacity in being able to view the entire duration of action clip.

\textbf{Latency Constraint:} The second major constraint for real-world skeleton-based action recognition is latency. While some existing works focus on the computational complexity of skeleton-based action recognition, many ignore the number of frames required to perform recognition of a single action. Typically, the amount of frames needed to classify a single action (i.e. clip size or $T$) of existing works is 300. If we are to assume a neural pose estimator running at $FPS_{in}=30$ (a non-trivial achievement depending on the hardware), then a 300 frame clip corresponds to a 10-second delay. This is prohibitively expensive for applications requiring real-time responses, such as patient monitoring, public safety, and general video surveillance.

\textbf{Edge Computing Constraint:} Lastly, an edge computing paradigm becomes a necessity due to the additional privacy and scalability considerations for real-world skeleton action recognition. Most existing video surveillance implementations stream image data from a local camera on-the-edge to cloud servers where the bulk of the computation would be performed. This invalidates user privacy and limits scalability. Privacy is essential toward enabling smart edge video applications to the point of other works dedicating computationally expensive deep neural networks \cite{taghavi2020edgemask} to insure user privacy.
In our work, low-power edge devices near the sensors are utilized for computing both the pose estimation and action recognition. By working directly on the video stream at the edge, the only data passed onto the network is pre-processed and contains no personally identifiable information. 

This inherently addresses the growing concerns of streaming personal data across the cloud. Additionally, the use of edge devices to handle the bulk of processing positively impacts the scalability of the system. In a cloud computing based system, scalability is limited by the processing power of the cloud server and the throughput of the network. Additional cameras lead to more requests for the server, which can lead to either costly server upgrades or an increase in system-wide latency due to the inability to meet system computation demand. When the system implementation is too costly it can limit the proliferation of smart video applications on-the-edge as well as the democratization of AI as a whole. Cutting back on costly cloud servers with cheaper local edge computation solves this issue, and has been used in existing deep neural network service platforms \cite{lu2020edgecompression,muksch2020benchmarking}. However, edge solutions are often tightly resource constrained, which results in other implementations utilizing hybrid edge/cloud solutions \cite{nigade2020clownfish}, or optimizing the DNN for performance \cite{qi2018enabling,nikouei2018real}.

\subsection{Real-World Evaluation Metrics}
In order to address latency constraints on resource constrained edge systems, we define two new metrics for analyzing and investigating the real-world deployment capability of action recognition algorithms. The first is Actions per Second $(ApS)$.

\begin{equation}
    ApS= \frac{T}{W}\times CPS_{in}
    \label{eq:ApS}
\end{equation}

This metric conveys how much work the model is able to complete in a single second (i.e, throughput). Where $T$ is the clip size, $W$ is the window size, and $CpS_{in}$ is clips per second or the rate a full clip is fed to the network. This metric is needed to compare latency constrained works such as ours to unconstrained works, as they do differing amounts of work in a single clip. While the theoretical input throughput ($FPS_{T}$), accepted by the model is important for fast moving applications such as self-driving cars, the domain of video surveillance has much more lax requirements (~30 $FPS$ for human motion). Instead, the second metric of Action Product Delay ($APD$), is much more predictive of the real-world deployment performance in domains such as video surveillance.

\begin{equation}
    APD= \frac{1}{FPS_{in}}\times W
    \label{eq:APD}
\end{equation}

$APD$ is the product between the inverse of the input frame rate $FPS_{in}$ and the window size of the network $W$. This metric conveys how much delay is needed to process a single action. In applications such as monitoring patients or hazardous working environments, the response time of detecting an incident (e.g. falling, tripping) is critical and its lower bound is the $APD$.

\subsection{End-to-End Video Surveillance and Noisy Data}

In a real-world setting, skeleton-based action recognition will be part of a larger end-to-end vision pipeline. \figref{fig:end-to-end} presents a typical vision pipeline for smart video surveillance applications. The figure is inspired by existing vision pipeline frameworks as presented in \cite{revamp2t} and \cite{MTADataset}. Shown in \figref{fig:end-to-end}, these frameworks generally consist of three conceptual stages. We also included an additional stage which consists of our RW-GCN network. A detailed description of RW-GCN can be found in \secref{sec:implementation}. The first is the detection stage. A multi-person human pose estimator is used to detect person keypoints and generate bounding boxes from a single frame of the input video. Next, in the extraction stage, person image crops are generated using the bounding boxes from the previous stage. These person crops are fed through a feature extractor network that generates an embedded feature representation for each person image. These bounding boxes are also used to calculate the temporal locality of each individual in the frame compared with previous frames. In the matching stage, a matching algorithm measures the euclidean appearance similarity between features from the current frame and those seen previously and, taking into account the previously calculated temporal locality, determines what IDs to assign to which person. 

This end-to-end vision pipeline is ran entirely on an edge node directly near the camera, with non-sensitive information aggregated from multiple nodes to an edge server, as done in \cite{revamp2t}.

\begin{figure}
    \centering
    \includegraphics[width=1\linewidth, trim= 18 15 18 15, clip]{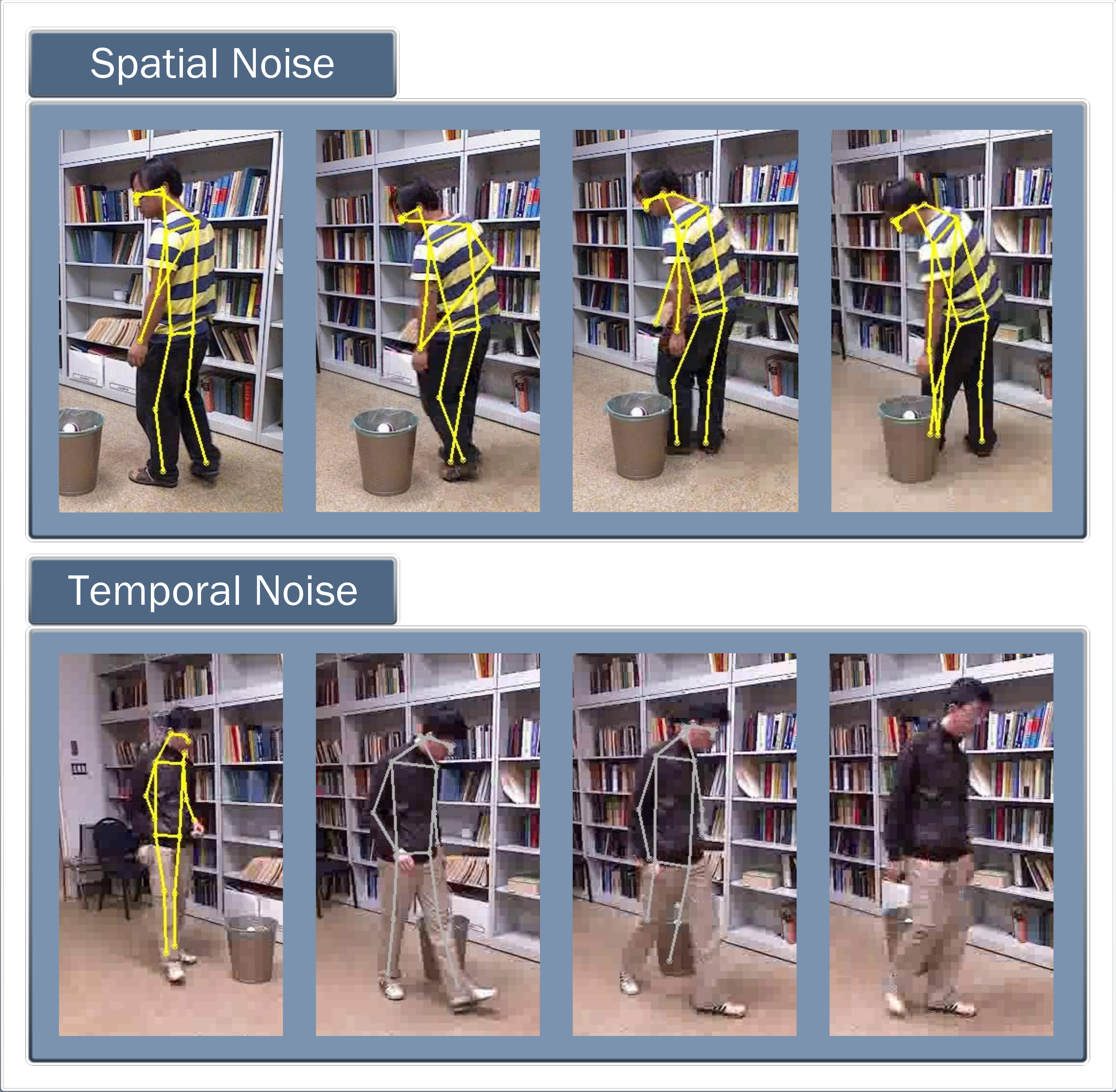}
    \caption{Visualization of spatial noise (missing keypoints) and temporal noise (missing identity). Grey indicates skeletons that have not been matched with an identity.}
    \label{fig:Noise}
\end{figure}

\textbf{Spatial-Noise:} Skeleton-based action recognition requires human pose information, in the form of a person keypoint skeleton, as an input. In closed-world action recognition, these skeletons are assumed to be flawless, perfect representations of the actors in the scene. However, in the real world, skeletons provided to the network by a human pose estimator will often be flawed. Even state-of-the-art methods are not perfect and will prove to be incorrect on occasion by either misdetecting a keypoint or failing to detect it all together. These types of errors may be inevitable on the whole, but they are even further exacerbated when the application environment differs from the dataset with which the pose estimator was trained, which will almost always be the case in the real world. For systems where hardware resources must be considered, or where real-time restraints exist, using the large, highly accurate state-of-the-art models may not be an option. In these cases a trade-off must be made, using lighter weight pose estimators and often sacrificing accuracy in order to meet application requirements, which can further increase the frequency of these errors. 

These missing and misdetected keypoints create a form of spatial noise in the system, as seen in \figref{fig:Noise}. This noise can be highly problematic when propagated to action recognition. The actor in the top half of \figref{fig:Noise} is leaning down to place an item in the bucket. This is clear to the human eye. However, the keypoint skeleton that is used for action recognition has prominent errors in this scene, particularly concerning the actors right arm. In addition to being generally inconsistent in its detection, in the far right image in the figure, the actors right hand is falsely detected all the way at his ankles. If an action recognition system is designed to assume the data it receives is perfect, this kind of error can easily confuse the network and lead to an incorrect action classification. As such, this type of noise needs to be accounted for in real-world action recognition. 

To balance between accuracy and real-time constraints, we use EfficientHRNet \cite{efficientHRNet} for acquiring human pose skeletons.

\textbf{Temporal Noise:} Another category of noise common to real-world action recognition is temporal noise. Temporal noise can occur in one of two ways. The first is when the skeleton for an actor is missing in one or more frames of a video segment. Generally this is either because the skeleton was not detected by the pose estimator, or because the actor's skeleton was not correctly matched to the actor for this frame. In both cases, this leads to the action recognition network having no skeleton information for the actor when such information should be present. This is not an issue in closed-world action recognition where perfect skeleton and ID data is provided, but in real-world action detection, where imperfect algorithms are used to generate keypoints and IDs, this type error is quite common.

The second form of temporal noise occurs when the edge system does not feed all the frames of the action into the network at the same time. While this is not an issue for closed-world action recognition, in the real world this can often encounter in applications where the latency constraints dictate that only a small number of frames can be processed at once.

\figref{fig:Noise} illustrates an example of the first category of temporal noise; missing keypoint skeletons. While the actor in the bottom half of \figref{fig:Noise} is correctly identified in the first frame, the middle two frames have greyed out keypoint skeletons, indicating that they were not correctly matched to the actor. This is an error caused by the re-identification algorithm. Further, in the last frame, no skeleton was detected at all; a failure of the neural pose estimator. With only a single frame worth of pose data to go on, the action recognition network will have no concept of the actor's motion and position through time. Intuitively, this will make action recognition for the scene incredibly challenging, and such instances must be accounted for when designing action recognition for the real world. 


We opt to use a similar Re-ID methodology as in \cite{revamp2t}, a real-world solution that does not rely on facial recognition.

\section{Real-World-GCNs (RW-GCNs) through Attentive Feedback Augmentation}

To meet the domain constraints of real-world skeleton action recognition, this section presents Real-World GCNs (RW-GCNs). 
In RW-GCN, we leverage a static graph with a pre-defined structure that mimics the COCO dataset keypoint format \cite{coco} for the spatial graph representation. The temporal component is made up of one-to-one connections between each vertex through adjacent frames. This spatio-temporal human pose graph is seen by the network as a tensor $\mathbf{X} \in \mathbb{R}^{N \times T \times C \times M}$ where $N$ is the number of nodes, $T$ is the number of frames, $C$ is the number of channels used , and $M$ is the max number of separate human poses in a single frame throughout the clip. Additionally, an adjacency matrix $\mathbf{A} \in \mathbb{R}^{N \times N}$ is used by the network to define convolution over the graph.

\label{sec:4}
\begin{figure*}[t!]
    \centering
    \includegraphics[width=1\linewidth, trim= 10 10 10 10, clip]{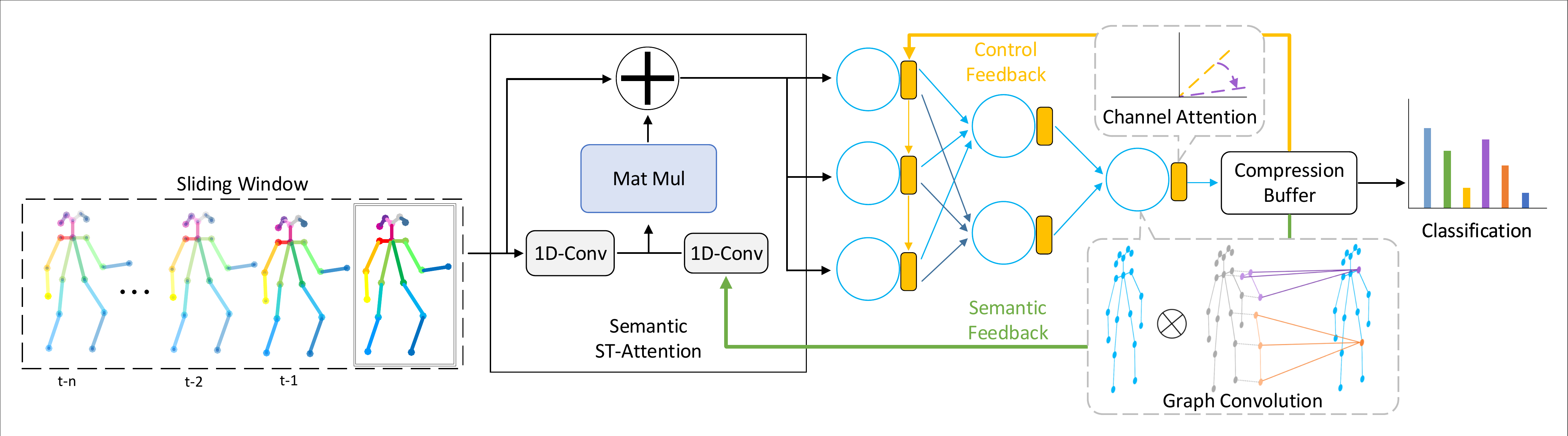}
    \caption{Feedback Augmented Spatial-Temporal Graph Convolution Network}
    \label{fig:RW-GCN}
\end{figure*}

\begin{equation}
\quad\quad\mathbf{Y}=\sum_{p \in \mathcal{P}} \overline{\mathbf{A}}_{P} \mathbf{X} \mathbf{W}_{P}
\label{eq:GCN}
\end{equation}

The spatial component of our ST-GCN based network is implemented by performing a $1 \times P$ standard 2D convolution on a single frame of the input buffer, where P dictates the number of partitions for the weight matrix and adjacency matrix. The result is then multiplied with a normalized set of $P$ adjacency matrices. Depending on the defined partitioning strategies, the weight matrix and the adjacency matrices maybe split into subsets and summed together as seen in equation \eqref{eq:GCN}, which is explained further in \cite{cheng2020skeleton}. RW-GCN leverages a spatial partitioning strategy as defined in \cite{yan2018spatial}.

RW-GCNs augment existing ST-GCN architectures with attentive feedback. Feedback can potentially enhance the capabilities of ST-GCNs in dealing with noisy inputs. The features feed back into the architecture will include higher level semantics, which may be redundant if there is a lack of change through time. These temporally redundant features can become spatially synergistic if the neural pose estimator's noise was correlated with its estimation, which we can qualitatively say is true. Overall, attentive feedback augmentation allows RW-GCNs to be applied to the domain of real-world skeleton action recognition by coping with the various noise, specifically the temporal noise induced by real-world latency constraints. As a result, RW-GCNs open up the possibility of skeleton-based action recognition applications for the edge computing domain. For systems where most of the computation is on a near sensor edge node, such as REVAMP\textsuperscript{2}T \cite{revamp2t}, RW-GCNs add a flexible network that can work with a varying number of frames depending on the computational intensity of the other networks. For systems such as EdgeEye \cite{liu2018edgeeye}, where all the data is sent to a localized edge server, RW-GCNs augment the edge system's flexibility in dealing with the network bandwidth and sensor storage limitations. Lastly, for more distributed edge systems such as VideoPipe \cite{salehe2019videopipe}, the flexibility of RW-GCNs with respect to local storage capacity and network bandwidth would enable many diverse pipeline implementations, each specialized towards the specific application.

\subsection{Theoretical Aspects}
The theoretical foundations for our attentive feedback augmentations are based on the information bottleneck theory of deep neural networks, proposed by Tishby et al. \cite{tishby2015deep}

While this theory was initially contested by the work of Saxe et al. \cite{saxe2019information}, the theory was then re-validated by the work of Noshad et al. \cite{scalable_noshad2019} stating that there was an issue in the calculation of mutual information due to poor mutual information estimators being used, with Nosahd et al. proposing their own Mutual information estimator. Additionally, further attempts have been made to verify the information bottleneck on more complicated architectures \cite{darlow2020information}. Despite the difficulties in directly applying the information bottleneck to deep learning, other works have shown clear empirical improvements in generalization when developing solutions based on the foundations of the information bottleneck \cite{jeon2021ib,perinlearning}. Our work takes the same approach, and develops architecture level solutions to real-world real-time action recognition using principles inspired from the information bottleneck theory of deep neural networks.

The information bottleneck theory starts with assuming the deep neural network can be represented as a Markov chain: 
\begin{equation*}
    X\rightarrow T_1\rightarrow ...\rightarrow T_z\rightarrow Y 
\end{equation*}

Each layer's output is represented by a random variable in the Markov chain. $X$ for example is the input feature, $T_1$ is layer $1$'s hidden representation, while $T_z$ is layer $Z$'s hidden representation and Y is the output random variable.

The information bottleneck when applied to deep learning states that learning is equivalent to minimizing the following Lagrangian:
\begin{equation}
    \label{info-lag-eq}
\min \; I(X;T)-\beta I(T;Y)
\end{equation}

We can interpret equation \eqref{info-lag-eq} as learning having two distinct and separate phases, by  minimizing $-I(T;Y)$, which we can treat as maximizing the inverse, and minimizing $I(X;T)$. The fitting stage is the first and is shown in equation \eqref{info-fit-eq}:
\begin{equation}
    \label{info-fit-eq}
\max \; I(T;Y) = H(T)-H(T|Y)
\end{equation}
The fitting phase is regulated by $\beta$, and learns informative features by maximizing the entropy of hidden features $H(T)$, which makes the hidden representation more informative. Additionally, the fitting phase minimizes $H(T|Y)$, which minimizes the information a hidden representation has when information on the label is already known. This can intuitively be seen as removing features irrelevant to the hidden representation.

The second phase of learning is the compression phase as shown in equation \eqref{info-comp-eq}:
 \begin{equation}
    \label{info-comp-eq}
\min \; I(X;T) = H(X)-H(X|T)
\end{equation}
The compression phase minimizes $H(X)$, the entropy on the source data, while maximizing the the amount of information a hidden representation contains about the input, $H(X|T)$. This leads to a model that realizes less on the input information. Our work hypothesizes that utilizing solutions with properties that assist either the fitting or the compression learning phases can lead to robust algorithmic solutions similar to Jeon et al. \cite{jeon2021ib}.

The first architecture level solution RW-GCNs employs is based on self-attention. The work of Bloem et al. \cite{probabilityinvariant_bloem2019probabilistic} states that attention allows networks to learn and model data-dependent invariance via capturing the minimum sufficient statistics of the data. Understanding that the information bottleneck can be seen as a framework for finding the minimal sufficient statistics~\cite{shwartz2017opening}, we hypothesise that attention enables models to better capture the minimum sufficient statistics of the dataset more efficiently, which directly helps the model with respect to the the information bottleneck framework. This is further supported by the perspective that attention can operate as a form of pre-training compression~\cite{probabilityinvariant_bloem2019probabilistic}. We additionally highlight the fact that early stopping (training only until validation accuracy stops increasing) is often employed in training deep neural networks. Other works have noticed that early stopping often (but not always) stops training before the compression phase, suggesting that the compression phase can lead to overfitting \cite{perinlearning,wickstrom2019information}. Knowing that, the compression phase can be interpreted as learning which information to focus on. We believe that self-attention is a flexible enough inductive bias (due to the ability to imitate and extend the inductive bias of convolution \cite{cordonnier2019relationship,bello2019attention}), to enable a non-overfitting compression phase.

Feedback can also assist model generalization, not through the compression phase but rather the fitting phase. In fact, the main premise of RW-GCNs is to maintain good representations or maximize $H(T)$ despite the presence of receptive field level temporal noise (induced by latency constraints). With that premise in mind, we look at the fitting phase, \textit{max} $I(T;Y)$, as a classical channel coding problem where our network is both the encoder and the channel itself and channel capacity is the mutual information $I(T;Y)$. We theorize that if feedback can assist channel encoding in classical communication channels, then it can assist the deep neural networks though improving the fitting phase of neural network training.

It is known that when feedback is applied to a traditional communication channel with uncorrelated Gaussian noise channel capacity does not increase~\cite{ebert1970capacity}. However, when the noise is correlated with the input (i.e. a channel with memory) feedback can lead to a small improvement in channel capacity. Alternatively, feedback can significantly reduce the complexity of the channel encoder and decoder \cite{chen2013variable} (i.e. the neural network), and through this reduced complexity the average block-length required to approach capacity in a traditional communication channel is reduced. We believe this is equivalent to reducing the amount of features needed to be seen by a deep neural network in a single time step. We relate this to how convolution neural network layers can be seen having neurons with a limited receptive field (i.e. less spatial features), and the network depth as mechanism that imitates feedback's ability to achieve reasonable accuracy with reduced complexity. It is known that depth can imitate feedback due to the fact that deep neural networks often have high amount of inter-layer weight sharing~\cite{boulch2017sharesnet,savarese2019learning}.

Feedback can additionally provide an additive gain in the capacity of memoryless Multi-Access Channels (MACs), where multiple signals are passed into the same channel \cite{gaarder1975capacity}. This can be seen as the neural network dealing with multiple features. The capacity improvement is achieved by feedback's ability to increase the cooperation of the signals, enabling the neural network to better jointly optimize multiple competing features. When only partial information of the multiple signals is desired, a multiplicative gain in channel capacity is achieved \cite{suh2011feedback}. 
We identify \textbf{semantic feedback} as a mechanism  that attempts to maximize $H(T)$, or equivalently increase the information of feature representations by exploiting the high level semantic information (such as deep neural network features), as seen in CliqueNets \cite{yang2018convolutional}. We believe this high level semantic information works through feedback's ability to reduce the complexity of the encoder (i.e. the network) which allows us to work with a smaller temporal receptive field.


Additionally, we identify \textbf{control feedback} as a mechanism that enables the extraction of statistics relevant to specific random variables but irrelevant to others. This type of feedback focuses on minimizing $H(T|Y)$, which reduces the redundancy of hidden features when some information about the target classification is known. This can be seen as an analog to feedback's ability to coordinate multiple signals in a traditional multi-access communication channel. Additionally, It is known that the information bottleneck leads to models with alternating formulations when conditionally optimizing to fit multiple random variables as seen in \cite{SIDEinfochechik2003extracting}. It is potentially possible to learn conditional models that focus on different output random variables depending on the feedback. Intuitively, if we assume a deep neural network model with the classes of tree, dog, and person, the best model would not expose the high-level features of a tree to the classification of the human class. Control feedback potentially can enable this conditional information flow.

\subsection{Implementation details} \label{sec:implementation}

In order to address the domain constraints discussed in \secref{sec:3}, we augment existing ST-GCNs with attentive feedback, creating Real-World Graph Convolutional Networks (RW-GCNs). In all variations of RW-GCNs we assume a latency constraint in the form of a window size $W$. When $W$ is equivalent to the total number of frames in an action clip $T$, RW-GCN functions as a general ST-GCN. For the baseline implementation of RW-GCN, a clip of $T$ frames is broken down into $W$ windows of $\frac{T}{W}$ frames, and an action classification is produced for each of those windows. RW-GCN adopts a sliding window operation with a stride equivalent to $\frac{T}{W}$. 

This limited window size leads to a limited temporal receptive field. To overcome this, RW-GCN propagates information from previous windows into the network while processing the current window. To this end, we explore two specialized forms of feedback, (1) semantic feedback and (2) control feedback for RW-GCNs. Each formulation works with the past feature as an input (in addition to current features) and implements two specialized forms of attention (semantic attention and channel attention). In the following, we discuss each in details.

\subsubsection{Semantic Feedback:} 
Our semantic feedback is designed with two goals in mind. The first is addressing both spatial and temporal noise while considering the latency constraint. This is achieved by first only operating on a window (or temporal slice) of the input and preserving the high-level semantic features from the last window for the processing of the next. We then feed these features into our Semantic Spatial-Temporal Attention block (semantic attention block) as shown in \figref{fig:RW-GCN}.

\begin{figure}[thb!]
    \centering
    \includegraphics[width=\linewidth, trim= 1 1 1 1, clip]{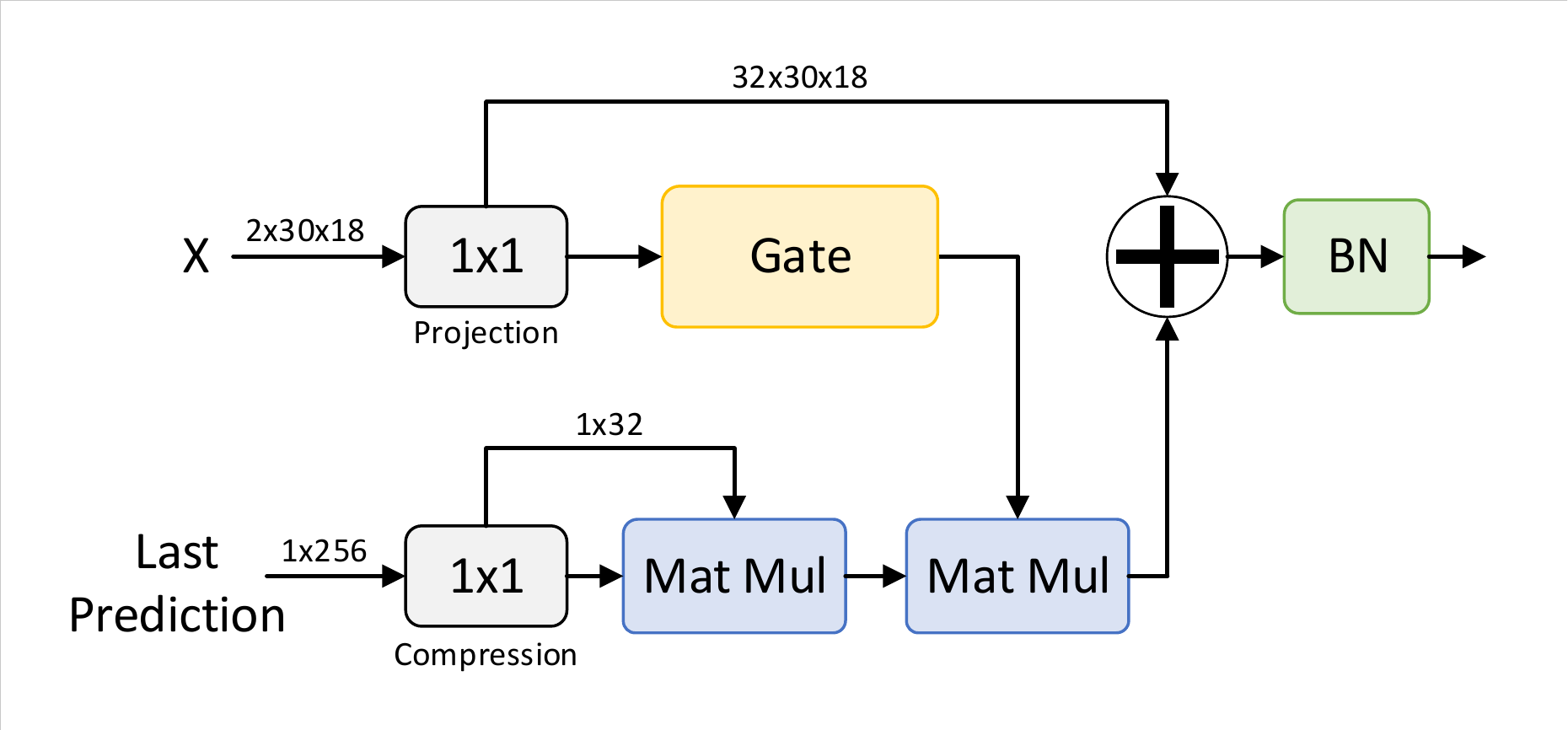}
    \caption{Feedback Augmented Attention}
    \label{fig:FB-ATT}
\end{figure}

The semantic attention block differs significantly from the original self-attention mechanism, which is typically in the form of equation \eqref{SA-eq}.

\begin{equation}
    \label{SA-eq}
    \text { Self-Attention }(Q, K, V)=\operatorname{softmax}\left(\frac{Q K^{T}}{\sqrt{d_{k}}}\right) V
\end{equation}

This implementation is achieved with a dot product between the past features and the current features rather than just the current features. This is a form of cross-attention. If the temporal information in $X$ is highly redundant, then our semantic attention formulation will be nearly equivalent to simplified self-attention \cite{simpleAttention}. This can give the model the opportunity to capture stronger spatial features, and theoretically deal with spatial noise. However, in the case of diverse temporal features (i.e. really fast motion), the semantic attention block enables the capture of pairwise interactions across frames, allowing for more refined feature representations through maximizing $H(T)$. 


Additionally, our semantic attention is designed to be computationally efficient. Simplifying the formulation of \eqref{SA-eq}, our semantic attention block combines key and value calculations into a single projection of $X$, instead of two \cite{guo2021attention,simpleAttention}. We also take inspiration from the work done in \cite{simpleAttention} and do not normalize the output. When we replace one of the inputs of $X$ with our past features $FB$, this enables our semantic attention block to capture semantic feedback.

To further enhance computational efficiency, we compress the past features into a $1X32$ tensor before applying semantic attention. This limits the computational costs despite the quadratic scaling compute complexity of attention mechanisms. Lastly, we add gates in the form of a $1 \times 1$ convolution followed by zero-initialized batch normalization, as well as residual with a multiplicative gate on the attention features. These gates help with the convergence of the network. The final formulation of our feedback augmented semantic attention block is seen in \figref{fig:FB-ATT}, and this formulation enables our ST-GCN to achieve semantic feedback.

The semantic attention block is placed in front of a standard ST-GCN, as shown in \figref{fig:RW-GCN}. In this work we keep the same underlying architecture of Yan et al. \cite{yan2018spatial} shown in \tabref{arch-tbl}. It is important to note that the ST-GCN block is composed of a GCN layer and a TCN layer. The GCN layer works as previously described in this section. The temporal convolution is simply a $1 \times 9$ convolution through the temporal dimension with a varying stride.

\begin{table}[t!]

    \centering
    \begin{tabular}{c|c|c|c}
        Layer Type & Stride & Rep & Filter Shape\\
        \hline
        \hline
        ST-GCN Block & s1  & x1 & 3x2x64 \\
        ST-GCN Block & s1 & x3 & 3x64x64 \\
        ST-GCN Block & s2 & x1 & 3x64x128 \\
        ST-GCN Block & s1 & x3 & 3x128x128 \\
        ST-GCN Block & s2 & x1 & 3x128x256 \\
        ST-GCN Block & s1 & x3 & 3x256x256 \\
        1D-GAP & N/A & x3  & N/A \\
        FCN & N/A & x1 & 256x120 \\
     
    \end{tabular}
    \caption{Standard architecture of ST-GCNs as seen in \cite{yan2018spatial}}
    \label{arch-tbl}
\end{table}

\subsubsection{Control Feedback}
Alternatively to implementing semantic feedback, RW-GCN can also implement control feedback. Control feedback only utilizes the compressed feedback features of the semantic attention formulation as shown in \figref{fig:FB-ATT}. Rather than modeling the pairwise interactions between features through time and space, our implementation of control feedback uses the past spatial-temporal features to actively weigh the channels, and effectively control which features propagate more strongly. We do this by utilizing a simple efficient channel attention \cite{wang2020eca}. This can be interpreted as feature selection and allows our network as a whole to theoretically alter its information flow with respect to the past signal in anticipation of the future signal (i.e. minimizing $H(T|Y)$ or equivalently reducing redundant features).

\section{Experimental Results and Evaluation}
\label{sec:5}
This work evaluates several variations of our proposed RW-GCN in the domain of real-world skeleton action recognition focusing on video surveillance applications. To this end, we distinguish three variations of RW-GCN: (1) Baseline, (2) controlled feedback, and (3) semantic feedback. The baseline implementation of RW-GCN does not utilize semantic or control feedback, and instead simply averages the past frame windows with the current one. We call this variation a consensus RW-GCN. In contrast, our semantic feedback variation is called RW-GCN-SF and our controlled feedback is called RW-GCN-CF.
All experiments were trained on a Nvidia-V100 GPU using Nesterov accelerated stochastic gradient descent with a momentum term of $M=.9$ and a weight decay set to $WD=10^{-4}$. Unless stated otherwise, the base starting learning rate is $lr=.01$, and has a decay of 10. Additionally, when training any feedback augmented RW-GCNs (not consensus), we implement a growing architecture strategy at certain epochs. We take the weights of a base RW-GCN and "grow" the semantic attention and efficient channel attention for either the semantic and control feedback. We believe these attention models can lead to a more robust compression phase by better capturing the invariance of the data \cite{probabilityinvariant_bloem2019probabilistic,wickstrom2019information}.

\subsection{Latency Constraint Evaluation and Analysis}

For the latency constrained accuracy experiments, we focus on the NTU-RGB-D-120 \cite{liu2020ntu} dataset. The NTU-RGB-D-120 dataset is the current largest Pose RGB+D human action recognition dataset. This dataset contains 120 different indoor action classes and is the current de facto standard for evaluating skeleton action recognition works. Each sample clip has a possible maximum of 300 frames at 30 FPS. Additionally, this dataset has a maximum of 2 people in a single frame concurrently. The dataset is composed of 106 unique subjects in combination with 32 possible setups. Overall it contains more than 114,000 video samples and 8 million frames. The original NTU-RGB-D-120 dataset has a pose format consisting of 25 nodes in 3D coordinate space $(x,y,z)$. We ignore the depth dimension and reduce the number of nodes to 18, in order to be compatible with end-to-end systems such as REVAMP\textsuperscript{2}T.

The specific hyperparameters of this training utilized a batch size of $N=32$. The baseline RW-GCN models are trained with a learning rate decay of $10^{-1}$ at epoch 30 and again at 60. For this dataset, we only pay attention to the first skeleton sequence even if there may be two actors in a scene. The RW-GCN-SF variation employs a learning rate restart at the 40th epoch and trains for 45 more epochs, for a total of 85 epochs. All consensus models were trained for 60 epochs. We do not test RW-GCN-CFs with this dataset. When validating our results we use the cross-setup and cross-subject validation strategies \cite{liu2020ntu}.

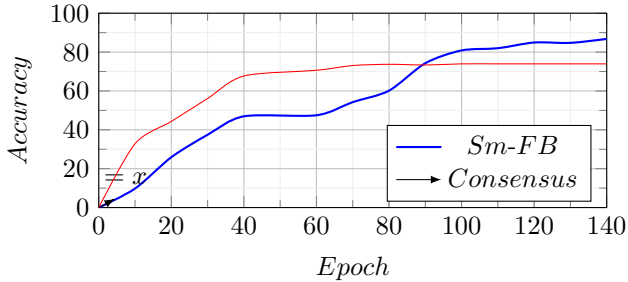
\begin{figure}[tb!]
\centering
\begin{tikzpicture}
    \begin{axis}[
        xmin = 0, xmax = 140,
        ymin = 0, ymax = 100,
        xtick={0,20,...,140},
        ytick={0,20,...,100},
        xtick distance = 2.5,
        ytick distance = 2.5,
        grid = both,
        legend pos=south east,
        minor tick num = 1,
        major grid style = {lightgray},
        minor grid style ={lightgray!25},
        width = \linewidth,
        height = 0.5\linewidth,
        ylabel = {$Accuracy$},
        xlabel = {$Epoch$},
        ylabel near ticks,]
 
    \addplot[ name path=f,
        smooth,
        thick,
        blue,
    ] file[skip first] {win30-cycle.dat};
    \addplot [name path = B,
        -latex,
        domain = 0:4.5] {x} 
        node [pos=1, above] {$y=x$};
         
    \addplot[
        smooth,
        red
    ] file[skip first] {win30-reg.dat};
    
    \addplot [teal!30] fill between [of = f and B, soft clip={domain=2:4}];
 
     \legend{$Sm$-$FB$,$Consensus$}
    \end{axis}

\end{tikzpicture}

\caption{ The learning curves of the baseline consensus model and our feedback augmented semantic attention. This is done with a latency constraint of 30.}
\label{fig:learningCurves}
\end{figure}

The first experimental results we analyze are the accuracy curves of training a baseline RW-GCN consensus model compared to a RW-GCN-SF, which utilizes semantic feedback. Due to the larger number of epochs used to train the feedback augmented networks, it is necessary to decouple the contributions of the training hyperparameters from the contributions of our attentive feedback augmentations. \figref{fig:learningCurves} shows when the training of the consensus model is extended to a similar number of epochs it stagnates at its existing accuracy. These learning curves are done with a learning rate restart at 50 and 90 epochs and without the growing training strategy. We find that when training with growing feedback augmented attention, RW-GCNs can reach similar accuracies, in fewer epochs, avoiding overfitting.

We go on to compare the accuracy of latency-constrained RW-GCNs to multiple existing works previously mentioned in \secref{sec:2}. This does not include works with extra modalities such as RGB patches. \tabref{tbl-ntu-lc-baseline} shows us that just by utilizing COCO formatted keypoints RW-GCN is able to achieve $87.7\%$ Cross-Subject Accuracy, a $16.4\%$ increase over the regular NTU-RGB-D skeleton baseline. This is done with no latency constraint and is only $1.6\%$ away from the previous SotA work DSTA-net \cite{shi2020decoupled}. We believe this increase in performance is due to the dimensional reduction of the input. By reducing the nodes from 25 to 18 and removing the depth channel, we reduce the dimensionality of the input by $48\%$. These dimensions can intuitively be explained as redundant and allow the model to focus on learning rather than dimension reduction and feature selection.

\begin{table}[htbp]
    \centering
    \resizebox{\linewidth}{!}{\begin{tabular}{c|c|c}
    
               & \multicolumn{2}{c}{Accuracy} \\
        Method & Cross-Setup & Cross-Subject\\
        \hline
        \hline
        ST-GCN \cite{yan2018spatial}\cite{papadopoulos2021vertex} & $71.3\%$ &$72.4\%$\\
        ST-TR-agcn \cite{li2019GEO}  & $84.7\%$  &82.7\%\\ 
        FGCN \cite{yang2020feedback}  & $87.4\%$  &85.4\%\\ 
        4s-Shift-GCN \cite{cheng2020skeleton}  & $87.6\%$&  85.9\%  \\ 
        ST-GCN-COCO (Ours) \cite{yan2018spatial} & \textbf{89.6}\%& $87.7\%$   \\
        PA-ResGCN-B19 \cite{shao2020MSNN} & $88.3\%$  &87.3\% \\
        MS-G3D Net \cite{liu2020disentangling} & $88.4\%$  &86.9\%  \\
        DSTA-net \cite{shi2020decoupled}  & $89.0\%$& $86.6\%$ \\ 
        RW-GCN (30 Frames)    & $73.3\%$  & 70.6\%  \\
        RW-GCN-SF (30 Frames) &80.1\%$\dagger$ & 86.4\%  \\
        RW-GCN-SF (100 Frames) & 89.5\%$\dagger$ & \textbf{94.16\%} \\
        
    \end{tabular}}
    \caption{Accuracy comparison against existing works on the NTU-RGB-D-120 action dataset. Cross setup validation leaves out half of the view points, while cross subject leaves out half of the subjects as defined in \cite{liu2020ntu}. $\dagger$ has 5 more training epochs.}
    \label{tbl-ntu-lc-baseline}
\end{table}

The Cross-Subject results go on to further show that our consensus-based RW-GCN with a latency constraint of $W=30$ performs at $70.6\%$, more than $17\%$ less than the SotA. This is done with $3\times$ fewer frames and therefore has $3 \times$ less delay between action classifications. In fact, assuming this work would be based on the 30 FPS input of the NTU-RGB-D-120 dataset, the SotA (and all existing works other than FGCN \cite{yang2020feedback}) would have an action product delay equating to a \textbf{10 second delay}, which can be prohibitively expensive for real-world applications such as video surveillance. RW-GCN has a \textbf{1 second delay}, which we classify as sub-real-time. When utilizing semantic feedback, RW-GCN-SF achieves $86.4\%$ accuracy, only $1.3\%$ away from the previous SotA. This significantly reduces the difference between the SotA while satisfying sub-real-time latency constraints, which enables emerging edge applications. When latency constraints are relaxed to allow a 3.33 second delay, we become the new SotA (with respect to Cross-Subject Validation) at \textbf{94.16\%} accuracy. This shows that not only does our method satisfy the sub-domain of Real-World Skeleton Action Recognition, but we make contributions to skeleton-based action recognition as a whole. When looking at Cross-Setup validation, we see our model does not define a new SotA but is only $.4\%$ less than the SotA. This discrepancy between Cross-Subject and Cross-Setup results can potentially be explained by two factors. The first is the lack of depth. Because of our reliance on COCO keypoints, we experience more domain shift between the training set and the validation set due to the differing viewpoints (and lack of features to inform the model of the sample viewpoints). The second factor is the temporal variance across subjects, which also causes a difference between validation and training set sample distributions. Our work has the potential to adapt to temporal variation more flexibly due to our attention feedback augmentations, and as such can learn to be invariant to individual subject temporal variances, minimizing the distribution shift between the training set and the validation set. This explanation is slightly contested by the baseline ST-GCN's better performance with Cross-Setup validation. This suggests that lack of depth is only an issue in the presence of temporal receptive field noise.

\begin{table}[htbp]
    \centering
    \begin{tabular}{c|c}
    
               & Accuracy \\
        Method & Cross-Subject \\
        \hline
        \hline
        Shift-GCN \cite{cheng2020skeleton} & $1.2\%$  \\
        ST-TR \cite{li2019GEO} & $5.9\%$  \\
        PA-ResGCN-B19 \cite{shao2020MSNN} & $10.12\%$  \\
        RW-GCN & $70.6\%$  \\
        RW-GCN-SF & \textbf{86.4\%}  \\
    \end{tabular}
    \caption{Accuracy comparison on the NTU-RGB-D-120 action dataset with real time latency constraints. Window size for all works is 30.}
    \label{tbl:lc-cross}
\end{table}

We believe this accuracy is achievable due to the semantic attention block's ability to encode spatial pairwise feature interaction when the temporal information is highly redundant or irrelevant (i.e. the frames within the window are minimally affected by receptive field noise), with sparse temporal feature interactions needed to encode the representation. As such, we believe our model encodes more discriminate spatial features than existing works when temporal noise is low, leading to our high accuracy. We additionally analyze other works' ability to match our performance with respect to latency constraints, shown in \tabref{tbl:lc-cross}. For this analysis we do not train the existing methods, but rather validate them with basic consensus feedback. We find that without training these works are unable to perform in the real-world domain as the best of them only achieved $10.12 \%$.

Our last latency constraint analysis on the NTU-RGB-D-120 dataset is an ablation study on generalization improvements of attentive semantic feedback augmentations to ST-GCNs over simple consensus-based RW-GCNs. This ablation allows us to see semantic feedback performing two modes of operation. In the first latency constraint of $W=30$, semantic feedback performs its intended functionality in allowing the model to retain past features through attentive pairwise feature interactions. The latency constraint of $W=30$ can potentially lack temporal redundancy and implicitly enforce a sparse sampling strategy similar to FGCNs \cite{yang2020feedback}. As a result, the model retains its ability to incorporate long-range temporal feature interactions and increases accuracy only $1.3\%$ off from its baseline unconstrained ST-GCN implementation, as seen in \tabref{tbl-ntu-lc-baseline}.

\definecolor{bblue}{HTML}{4F81BD}
\definecolor{ppurple}{HTML}{9F4C7C}
\begin{figure}[h]
    \centering
    \begin{tikzpicture}
        \begin{axis}[
            width  = .99\linewidth,
            height = 4cm,
            major x tick style = transparent,
            ybar=\pgflinewidth,
            bar width=12pt,
            ymajorgrids = true,
            ylabel = {Accuracy},
            xlabel = {Latency constraints},
            symbolic x coords={LC-30,LC-60,LC-100},
            xtick = data,
            scaled y ticks = false,
            enlarge x limits=0.25,
            ymin=50,
            ymax= 110,
            nodes near coords,
            point meta=y, 
            legend cell align=left,
            legend style={
                    at={(1,1.05)},
                    anchor=south east,
                    column sep=1ex
            }
        ]
            \addplot[style={bblue,fill=bblue,mark=none}]
                coordinates {(LC-30, 70.6) (LC-60,83.0) (LC-100,82.5)};

            \addplot[style={ppurple,fill=ppurple,mark=none}]
                coordinates {(LC-30, 86.4) (LC-60,89.4) (LC-100,94.16)};

            \legend{RW-GCN,RW-GCN-SF}
        \end{axis}
    \end{tikzpicture}
    \caption{ The ablation of latency constrained consensus and semantic feedback models for multiple constraints.}
    \label{fig:lc-varied}
\end{figure}
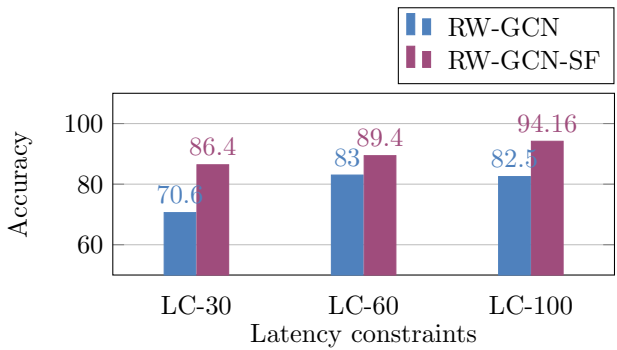

The mode of operation of the ST-GCN begins to shift when constrained by a window of $W=60$. The features it records are more temporally redundant, and the model naturally has the capacity to store longer-range temporal interactions. As such, the model improves the unconstrained baseline of $+2.3\%$. We believe that this window constraint is balancing improving its spatial discriminative power with maintaining the temporal feature retaining capabilities of the lingering implicit receptive field level temporal noise. Due to the complexity of these two tasks, the accuracy improvement over the consensus model is only $6.4\%$ as opposed to the $16.4\%$ improvement in the $W=30$ test. Lastly, with a constraint of $W=100$, the semantic attention block can maintain a minimum amount of temporal features representational power and focus on more competently encoding the spatial features. Due to the consensus model's inability to improve the spatial representational power of the ST-GCN, the consensus model's accuracy stagnates.

\subsection{End-to-End System Error Propagation Analysis}

The end-to-end system noise analysis goes beyond the latency constraint analysis and looks at the emergent system noises of a realistic end-to-end edge system. The Northwestern-UCLA dataset \cite{wang2014NwUCLA} is the data set used and is also a Pose-RGB-D human action recognition dataset. It contains 1494 videos. Each clip contains an average of 18.5 frames with a standard deviation of $\pm13.25$. The dataset has 10 possible action classes that are exceptionally suited to video surveillance applications. We utilize a Cross-View validation strategy employed by \cite{yang2020feedback}, where we leave out the last camera view.

In order to truly evaluate the end-to-end system noise applicable to video surveillance applications, we run the RGB images of this dataset though REVAMP\textsuperscript{2}T, with EfficientHRNet-H\textsubscript{0} \cite{efficientHRNet} and the MobileNetV2 \cite{mobilenets} feature extractor used in the original REVAMP\textsuperscript{2}T paper \cite{revamp2t}. This produces spatial noise in the keypoints due to the neural pose estimator and frame-level temporal noise brought by the Re-ID model, in addition to an extreme latency constraint of $W=10$.

Due to Re-ID varying the number of people detected and the natural temporal variance of the dataset, the input tensor $\mathbf{X} \in \mathbb{R}^{N \times T \times C \times M}$ has widely varying $T$ and $M$ dimensions. We utilize a dynamic batch preprocessing step to deal with this which first pads the $T$ dimension with repeating clips until it is the length of the largest clip in the batch. The $M$ dimension of the tensor represents multiple people. Due to the error prorogation of Re-ID into action recognition, we potentially have single-person clips displaced throughout the $M$ dimension in a disjoint and irregular manner. To solve this the dynamic batching will adaptively sum the input to a smaller $M=\lambda$. If $\lambda=1$, we call this the ID agnostic scenario where we are no longer modeling the error propagation from Re-ID. In the case of a sample having $M<\lambda$ we zero pad the tensor until $M=\lambda$.

For the training, all consensus models were trained for 30 epochs initially then go onto 60 epochs with a standard learning rate decay, while the feedback models are added on top of the consensus model's weights at epoch 30 and are trained for an additional 40 epochs. Due to the temporal noise injected by dynamic batching, we utilize a batch size of $8$. The validation batch size is set to 1 to minimize the temporal noise brought by dynamic batching. This set of experiments additionally utilizes a learnable multiplicative gate attached to each FB-Augmented attention, initialized at zero. This enables the feedback augmented models to be integrated into the consensus models smoothly.

\begin{table}[tbp]
\centering
    \begin{tabular}{c|c}
    
         Method & Accuracy \\
        \hline
        \hline
        RW-GCN   & $88.5\%$  \\ 
        HiGCN \cite{huang2019hierarchical} & $88.9\%$  \\ 
        Ensemble TS-LSTM \cite{lee2017ensemble} & $89.2\%$  \\ 
        MSNN \cite{shao2020MSNN} & $89.4\%$  \\
        FO-GASTM \cite{li2019GEO}  & $91.3\%$  \\ 
        Where to focus? \cite{das2019focus}   & $91.3\%$  \\ 
    
        AGC-LSTM \cite{si2019attentionLSTMGCN}  & $93.3\%$  \\ 
        4s-Shift-GCN \cite{si2019attentionLSTMGCN}  & $94.6\%$  \\ 
        FGCN \cite{yang2020feedback}  & 95.3\%  \\ 
        RW-GCN-SF & \textbf{96.6\%}  \\ 
        
        \hline
        \hline
        \multicolumn{2}{c}{Noisy (Ours)} \\
        \hline
        RW-GCN       & $84.9\%$  \\
        RW-GCN-CF    & $83.4\%$     \\
        RW-GCN-SF    & $90.4\%$     \\
        RW-GCN-SF+CF & $82.9\%$     \\  
    
    \end{tabular}
    \caption{Accuracy comparison with existing methods on the NW-UCLA action dataset in the real-world domain.}
    \label{nw-ucla-t1}
    \vspace{-20pt}
\end{table}

\tabref{nw-ucla-t1} shows the accuracy results over noisy-input on the NW-UCLA action dataset. It also compares our approach against FGCN \cite{yang2020feedback}, an existing work that also incorporates feedback, as the SotA. We surpass SotA by $1.3\%$ in the absence of both spatial keypoint noise and Re-ID noise. Furthermore, the results show that our RW-GCN-SF is able to contend with the FGCN architecture despite incorporating a latency constraint of $W=10$, noisy keypoints from EfficientHRNet-H0, and frame-level temporal noise due to the REVAMP\textsuperscript{2}T Re-ID. We report $90.4\%$ accuracy, only $4.9\%$ away from FGCN's SotA. Additionally, Yang et al. \cite{yang2020feedback} reports their accuracy with what we would consider a latency constraint of 64 frames (as opposed to our 10 frames), from which they focus on randomly sampling. 

We believe that our ability to achieve over $90\%$ accuracy despite the aforementioned noise is critical to real-world skeleton action recognition and smart edge video surveillance applications. This is further reinforced by our results in \tabref{tbl:lc-cross}, which demonstrate the inability of existing works to adapt to the receptive field level noise of latency constraints. Additionally, we go on to analyze the difference in accuracy between control feedback and semantic feedback models. \tabref{tbl-ntu-lc-baseline} highlights that control feedback performs worse than even the baseline consensus model, and fails to work with semantic feedback. Our primary hypothesis is that control feedback failed to perform as effectively as semantic feedback due to RW-GCN-SF modeling pairwise bi-linear feature interactions, while the RW-GCN-CF variant only modeled linear interactions. This is further exacerbated by the lack of temporal redundancy in the features due to the extreme latency constraint. Furthermore, when analyzing temporal noise injected by dynamic batching (by setting the validation batch $N>1$) control feedback methods achieved $84\%$, as opposed to semantic feedback's $83.1\%$. Our logical conjecture is that the irregular replication of features, through the dynamic batching process, imitates a noisy recurrent feedback where the same features are pushed through the same network weights repeatedly, and that acts as a form of information processing. This form of information processing is more akin to how traditional recurrent neural networks processed temporal information. It could be said our RW-GCNs exist in a family of models where the reliance of network depth and recurrent processing differs for each variation. The AGC-LSTM \cite{si2019attentionLSTMGCN} model would be a variation that is the opposite of our model with a heavy reliance on recurrence and attention, but little reliance on network depth. Control feedback may be more reliable than semantic feedback for models such as AGC-LSTMS, as it controls information propagation.  

\begin{table}[tbp]
    \centering
    \begin{tabular}{c|c|c|c}

                     & ID              & \multicolumn{2}{c}{Re-ID} \\
        Method       & Agnostic        & Noise 1   & Noise 2       \\
        \hline
        \hline
        RW-GCN       & $84.9$\%        &  \textbf{67.2\%} & $61.6$\% \\
        RW-GCN-SF    & \textbf{90.4\%} & $52.21$\% & \textbf{71.8}\% \\
        RW-GCN-CF    & $83.4$\%        & $40.9$\%  & $30.9$\% \\
        RW-GCN-SF+CF & $82.9\%$        & $64.60$\% & $36.9$\% \\
            
    \end{tabular}
    \caption{Ablation on Re-ID noise against network scenarios. Noise 1 is appearance centric feature matching based Re-ID, while Noise 2 is IoU temporal locality based Re-ID.}
    \label{tbl:noise-ablation}
    \vspace{-20pt}
\end{table}

The last analysis of the Northwestern-UCLA dataset \cite{wang2014NwUCLA} is an ablation study on the effect of Re-ID error propagation. We test RW-GCN architectures in two scenarios. The first is an ID agnostic scenario where the RW-GCN attempts to artificially remove temporal noise by summing across the $M$ dimension of the input to rejoin the displaced temporal features into one single $1\times T\times V\times C$ tensor. While this would theoretically test poorly on crowded scenes with multiple people performing separate actions, the Northwestern-UCLA dataset \cite{wang2014NwUCLA} has very few such samples. We see our best performance with the RW-GCN-SF work achieving $90.4\%$ action recognition accuracy. In the second scenario, we analyze real-world action recognition with respect to Re-ID noise. We vary the hyperparameters that weigh the temporal IoU filter against the appearance feature distance. We find that RW-GCN-SF is the most resilient at $71.8\%$ accuracy. An interesting aspect of our results shows that the appearance centric Noise 1 is overall more inhibiting to RW-GCNs. While this may be just an aspect of the dataset, we believe that this merits future exploration. 

Overall, we conclude that the zero-padding of the input feature along the $M$ dimension results in too much noise, which brings down the accuracy of all models in the real-world scenario. It was expected that IoU based Re-ID would outperform appearance-focused Re-ID, however, marginal improvements are achieved from the top network in the appearance-based noise test.

\subsection{Real-Time Performance Analysis}

The cross-analysis of our on-the-edge real-time performance metrics is shown in \tabref{tab:real-time}. The results were measured on a real-world edge device, the Nvidia Jetson Xavier NX. In order to fairly evaluate real-time performance, two metrics were introduced in \secref{sec:3}; Actions per Second ($ApS$), calculated by equation \eqref{eq:ApS}, and Action Product Delay ($APD$), calculated by equation \eqref{eq:APD}. It is important to note that $ApS$ is device specific, while $APD$ is end-to-end system specific.

\begin{table}[h]
    \centering
    \vspace{+3pt}
    \resizebox{\linewidth}{!}{
        \begin{tabular}{c|c|c|c|c|c}
            Method & Window & Acc      & Params &  ApS    & APD       \\
                   & Size   & (\%)     &  (M)   &         & (sec)      \\
            \hline
            \hline  
            S-TR \cite{li2019GEO}               & 300  & 84.7\%  & 3.07  & $12.7$  & $13$ \\
            Shift-GCN \cite{cheng2020skeleton}  & 300  & 87.8\%  & 0.77  & $25.1$  & $13$ \\ 
            ResGCN-B19 \cite{song2020stronger}  & 300  & 87.3\%  & 3.26  & $23.2$ & $13$ \\
            \hline
            RW-GCN-SF$\dagger$    & 10   & 71.8\%  & 3.22  & $15.6$  & $0.4$ \\
            RW-GCN       & 30   & 70.6\%  & 3.22  & $41.7$  & $1.3$ \\
            RW-GCN-SF    & 30   & 86.6\%  & 3.22  & $41.2$  & $1.3$ \\
            RW-GCN-SF    & 100   & 90.4\%  & 3.22  & $16.5$  & $4.3$ \\    

        \end{tabular}
    }
    \caption{Real-world comparison with state-of-the-art methods on the Nvidia Jetson Xavier NX. Actions per Second ($ApS$) measures classification throughput. Action Product Delay ($APD$) measures the latency of classification and was calculated with respect to EfficientHRNet-H\textsubscript{0}'s FPS. $\dagger$ This work reports accuracy on the Northwestern-UCLA dataset as opposed to NTU-RGB-D-120.}
    \label{tab:real-time}

\end{table}

Our model intuitively supports higher $ApS$ as it produces multiple action classifications in a clip. While some may disparage the necessity of multiple outputs per clip when considering a real-world scenario, it is important to acknowledge that the speed at which actions are performed is both unknown and data dependent. Additionally, action recognition models are typically not aware of the input frame rate (unless trained to be \cite{feichtenhofer2019slowfast}, which most works are not). This leads to the necessity of supporting larger $ApS$ than needed in ideal training scenarios, and provides future potential to train a model to be agnostic to varying input frame rates (which, when left unchecked, could lead to accuracy degradation). RW-GCN reports the highest performance at $41.7$ $ApS$, which is $1.6\times$ higher than existing works. It is seen that the overhead of semantic feedback in terms of $ApS$ and parameters is minimal. As mentioned in \secref{sec:4} semantic feedback was designed to be lightweight, only working on 32 channels. Shift-GCN is the most parameter efficient model as it was explicitly designed for lightweight and efficient inference, by utilizing data movement across time steps. This model design is not mutually exclusive with RW-GCNs, and in the future can be explored.

It is important to note that the $ApS$ of RW-GCN models varies wildly with respect to frame size. This is due to the execution bias of GPU-based systems. As we decrease window size, the data-dependent nature of RW-GCNs serializes and losses performance efficiency. On the other end of the spectrum, when the window size is increased we end up with the same problem as other works due to needing more frames to compute a single action. The $APD$ calculation is dependent on EfficientHRNet's H\textsubscript{0} FPS of 22.95 \cite{efficientHRNet}. RW-GCN achieves the lowest $APD$ of $0.4$ seconds with our extreme latency result derived from the Northwestern-UCLA dataset \cite{wang2014NwUCLA}.

\begin{figure}[h]
    \centering
    \begin{tikzpicture}
        \begin{axis}[
            width  = .8\linewidth,
            height = 4.2cm,
            major x tick style = transparent,
            ybar=\pgflinewidth,
            bar width=12pt,
            ymajorgrids = true,
            ylabel = {Actions per Second},
            xlabel = {Number of People},
            symbolic x coords={2,4,6,8},
            xtick = data,
            scaled y ticks = false,
            enlarge x limits=0.25,
            ymin=0,
            ymax= 18,
            nodes near coords,
            point meta=y, 
            legend cell align=left,
            legend style={
                    at={(1,1.05)},
                    anchor=south east,
                    column sep=1ex
            }
        ]
            \addplot[style={bblue,fill=bblue,mark=none}]
                coordinates {(2, 15.6) (4,11.0) (6,6.5) (8,5.5)};

        \end{axis}
    \end{tikzpicture}
    \caption{ Scalability analysis of RW-GCNs on Nvidia Jetson Nano}
    \label{fig:nano-APS}
\end{figure}
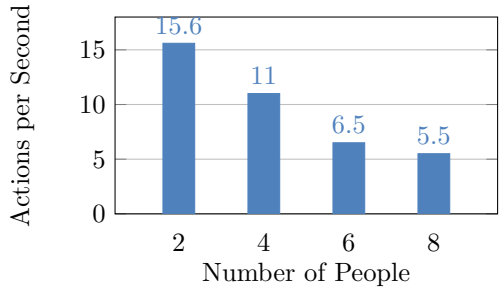

Lastly, we analyse the scalability of RW-GCNs on lightweight and inexpensive edge devices. We focus on the Nvidia Jetson Nano as opposed to the Nvidia Xavier NX, which was previously used for the results shown in \tabref{tab:real-time}. The Jetson Nano is nearly $10\times$ less expensive than the Xavier NX as of the writing of this paper. The price is an important factor when determining the scalability with respect to system costs, and in fact democratizing AI through affordability. However, the Nano is much more tightly resource constrained, with only a 128-core Maxwell GPU compared to the Xavier NX's 384-core Volta GPU and 48 Tensor cores. This is in addition to the greater RAM bandwidth ($\approx2\times$), and greater memory capacity ($\approx4\times$).  We use $ApS$ to measure how our compute maps to such a constrained environment. We calculate $ApS$ with a clip size of 30 and a window size of 10. We see a $2.67\times$ drop in $ApS$ with only two people in a single scene. This deficit increases to $7.6\times$ less $ApS$ when scaled to 8 people. We believe this to be more than adequate for real time action detection, and in fact believe that further trade offs of $ApS$ are exploitable for supporting scenes with larger amounts of people. Theoretically, we can sequentially execute across the number of people similar to our sliding window solution to large temporal frames. This could potentially allow us to process scenes with 36 people with an ApS of $1$. Future directions can explore further optimization through either offloading deeper layers to a server similar to \cite{nikouei2019safe}. By choosing to offload deep layers privacy is preserved. Another direction is utilizing input adaptive compute, similar to the work down by Fang et al. \cite{fang2020flexdnn}.

\section{Conclusion}
\label{sec:6}
In conclusion, this article defined the sub-domain of real-world skeleton action recognition and utilizing information theory based principles, we introduced Real-World Graph Convolutional Networks (RW-GCNs) to achieve a new SotA accuracy of \textbf{94.16}\% for NTU-RGB-D-120 that has $3.02\times$ less latency over the ST-GCN baseline.
Additionally, RW-GCNs can achieve 10X less latency over the baseline implementations with only 3.8\% less accuracy from our new SotA. Evaluating the Northwestern UCLA dataset \cite{wang2014NwUCLA} reveals that RW-GCNs can achieve 90.4\% accuracy with \textbf{32.5}$\times$ less latency than the baseline ST-GCNs. This is despite spatial keypoint noise being present in validation and training. Finally, RW-GCNs can run in the presence of fully end-to-end system noise, including temporal Re-ID noise, with $32.5\times$ less latency than baseline St-GCNs and maintaining $71.8$\% accuracy on the Northwestern UCLA dataset. This is all achieved with a privacy perceiving and scalable edge-computing centric methodology, where system cost can be reduced by \textbf{10}$\times$ per each node while still maintaining a respectful range of throughput dependent on scene complexity (~$15.6$ to $5.5$ $ApS$). 

This work marks the inception of the sub-domain of real-world skeleton action recognition. By devising RW-GCNs, this work hopes to facilitate the design and creation of new edge computing applications that were previously infeasible. However, we believe there still exist many challenges for this emergent domain, such as temporal variation, environment specific scene dynamics, and further application-specific constraints.

\section*{Acknowledgement}
This research is supported by the National Science Foundation (NSF) under Award No. 1831795.

\bibliographystyle{acm}
\bibliography{ref.bib}

\end{document}